\pdfoutput=1
\documentclass[conference]{IEEEtran}
\usepackage[utf8]{inputenc}
\usepackage{pgfplots}
\usepackage{pgfplotstable}
\pgfplotsset{compat=newest}
\usepgfplotslibrary{statistics}
\usepackage{tikz}

\usepackage{ifpdf}
\usepackage{cite}
\ifCLASSINFOpdf
\else
\fi
\usepackage{amsmath}
\usepackage{bm}
\usepackage{amssymb}
\usepackage{algorithm2e}
\usepackage{array}
\ifCLASSOPTIONcompsoc
 \usepackage[caption=false,font=normalsize,labelfont=sf,textfont=sf]{subfig}
\else
 \usepackage[caption=false,font=footnotesize]{subfig}
\fi
\usepackage{stfloats}
\usepackage{url}
\begin{document}
\title{Layerwise Noise Maximisation\\ to Train Low-Energy Deep Neural Networks}

%DONE: ***Hide author names for submission***
\author{
\IEEEauthorblockN{Sébastien Henwood, François Leduc-Primeau, and Yvon Savaria}
\IEEEauthorblockA{Department of Electrical Engineering, École Polytechnique de Montréal\\
Montréal (QC) Canada\\
Email: \{firstname.lastname\}@polymtl.ca}
}


%Telephone: (800) 555--1212, Fax: (888) 555--1212}


\maketitle

\begin{abstract}
Deep neural networks (DNNs) depend on the storage of a large number of parameters, which consumes an important portion of the energy used during inference.
This paper considers the case where the energy usage of memory elements can be reduced at the cost of reduced reliability. A training algorithm is proposed to optimize the reliability of the storage separately for each layer of the network, while incurring a negligible complexity overhead compared to a conventional stochastic gradient descent training. For an exponential energy-reliability model, the proposed training approach can decrease the memory energy consumption of a DNN with binary parameters by 3.3$\times$ at isoaccuracy, compared to a reliable implementation.
%We propose the Layerwise Noise Maximisation (LaNMax) algorithm to optimise a neural network's memory energy consumption. It manages to lower the energy required for a binary neural network by three times at isoaccuracy. It follows a SGD-like training in order to find an optimal trade-off between energy and accuracy given a parameter $\alpha$. This SGD is adapted such that estimates of the gradient are obtained by introducing small deviations on the current fault rate $\bm{p}$ at each mini-batch. There is little to none overhead to the LaNMax augmented training. This paper aims at bolstering the interest for deep learning techniques in IoT/embedded uses cases by allowing small energy footprints network to be efficiently trained.
\end{abstract}


% \fi
\IEEEpeerreviewmaketitle

\section{Introduction}

% \begin{itemize}
%     \item it is important to reduce the energy consumption of the inference phase of DL
%     \item the memory required for storing network parameters consumes an important part of the energy
%     \item naturally, diminishing the network size
%     \item furthermore, tinker with the memory energy supply
%     \item memory fault model (BSC, p)
%     \item traditional approach, shared $p$
%     \item lanmax to push the boundary
% \end{itemize}

Deep learning \cite{Lecun2015DeepLearning} has attracted a lot of interest since 2012 and AlexNet's achievements on ImageNet \cite{Krizhevsky2012ImageNetNetworks}, sparking a rapid improvement of the state-of-the-art in diverse fields. However the trade-off for such results is an antagonizing growth in resource requirements. 
%While they are bound to become ubiquitous, 
% We now see neural networks such as GPT-2\cite{Radford2018LanguageLearners} with an astonishing 1.5 billion parameters, resulting in important resource requirements during inference. 
We now see neural networks such as GPT-2~\cite{Radford2018LanguageLearners} with as much as 1.5 billion parameters.
%with adapted resources in storage, energy, compute inference wise, most of the regular computers can't handle easily such networks. 
%As the exponential trend continue, 
Such resource requirements make it difficult to implement high performance networks directly in IoT/embedded devices and imply a large energy usage. The energy consumed by memory cells can represent up to 60\% of the overall energy consumption\cite{Kim2018Energy-EfficientErrors}.
One natural countermeasure is to forsake a bit of accuracy and downscale the network 
thus reducing the number of parameters and operations. Another approach is to reduce the precision used to store the parameters: numerous quantization schemes have been proposed such as Binary Connect \cite{Courbariaux2015Binaryconnect:Propagations} (BC) that represents parameters using only 1 bit. 
Quantizing the parameters can reduce energy consumption at isoaccuracy, as shown for instance in \cite{Moons2018MinimumNetworks}. Hence, we consider networks binarized with BC as a starting solution on which we want to improve the energy consumption at isoaccuracy.

%DONE
% contexte, objectifs, travaux existants, ce que l'on fait en plus, résultats
% Offering neural networks reliability to faults has been a growing concern with the equal growth of computationnal requirements to run them. As noted in \cite{Torres-Huitzil2017FaultReview}, the intrinsic fault tolerance of neural networks could bolster the success of embedded applications. However, most of the work to increase the reliability of neural networks is focused on errors rather than faults, on results rather than low level logical gates. 

To reach that goal, the supply voltage of memory cells can be reduced to decrease the energy consumption. However, this can result in a higher likelihood of faults occurring at write or read time and causing data corruption.
%It however induces faults assumed to happen at a uniform rate $p$ when the memory is accessed.
Several previous works have proposed accepting memory faults to reduce the energy consumption of DNNs.
%The mechanism of a faulty memory has already been proposed to achieve more efficient networks, with a unique global fault rate $p$. 
A possible approach is to circumvent faults during inference using advanced fault detection schemes \cite{Reagen2016Minerva:Accelerators}. Sustaining faults through adapted training is also an option: for example, \cite{Hirtzlin2019OutstandingNetworks} explored the use of a faulty memory on a fully binary neural network and reported a tolerance to a fault rate up to 4\% of the memory cells. 
A training approach for networks with higher precision weights has also been proposed\cite{Hacene2019TrainingRobustness}, making use of fault detection schemes available in hardware. 

% FAULTY IN GENERAL
Another approach for energy savings in neural networks is to supply less energy to the computation units instead of the memory units, as for example in \cite{Shin2019SensitivityAccelerators}. While the assumed main type of error is slightly different (timing errors), the analysis is still focused on which weights are the most sensitive: authors propose to decide for the relative importance of weights by using a Taylor expansion of the computations. 

As suggested in \cite{Hacene2019TrainingRobustness}, the memory fault rate can be further optimized per block of a ResNet-style neural network. This echoes with the work from \cite{Zhang2019AreEqual}, in which a layerwise sensitivity analysis is proposed. They find that while some layers may be critical for the network accuracy, others could be randomized without too much performance degradation. 
However, optimizing the fault rate for each layer is not a straightforward task since faults occurring in each layer jointly affect the final accuracy.
We consider the case where no fault-detection mechanism is available, and consider the problem of optimizing the memory fault rate separately for each layer of the neural network to create more possibilities for energy reduction.
For this, we propose the Layerwise Noise Maximisation (LaNMax) algorithm to optimize the energy-accuracy tradeoff of the network automatically. Compared to a binarized neural network implemented using reliable memories, the LaNMax algorithm finds a fault-rate assignment that reduces memory energy by $3.3\times$ on the CIFAR10 dataset\cite{Krizhevsky2009LearningImages}, or $2.4\times$ when compared to a uniform-noise approach, at isoaccuracy.

The outline of this paper is as follows: Section~\ref{sec:memorymodel} introduces the memory fault model used in our experiments, Section~\ref{sec:alg} describes the LaNMax algorithm, Section~\ref{sec:exp} presents the experimental results obtained, and Section~\ref{sec:ccl} concludes the paper.

%%%%%%%%%%%%%%%%%%%%%%%%%%%%%%%%%%%%%%%%%%%%%%%%%%%
\section{Memory-fault model}\label{sec:memorymodel}
In this work, we consider networks binarized with BC: weights $\theta$ are taken in the set $\{-1,1\}$. It is assumed that supplying less voltage to the memory cells will create faults at rate $p$ when the parameters will be read, with an energy consumption $\eta$ following the formula
\begin{equation}\label{eq:conso}
    \eta(p) =  -\frac{\ln(p)}{a} \, ,
\end{equation}
where $a$ is a technology dependant parameter. 
Such an exponential energy-reliability tradeoff is observed for instance in CMOS SRAM circuits\cite{Dreslinski2010Near-ThresholdCircuits}.
A one-bit parameter $\tilde{\theta}$ read back from the faulty memory can be described with a binary symmetric channel (BSC) model, defined as
\begin{equation}
    BSC(\theta, p) = \begin{cases}
        -\theta & \text{with probability $p$,}\\
        \theta & \text{with probability $1-p$.}
    \end{cases}
\end{equation}
% . We denote this BSC by $BSC(\theta,p)$. 
We further assume that stored bits are affected independently by faults, which can be approximated in practice even in the presence of correlated bit-cell faults through interleaving. 

Since the memory can be modeled as a communication channel, we refer to $p$ as the noise level.
%\francois{[What are the underlying assumptions that allow us to use a BSC model? We assume that bit cells are affected independently by faults. However, unlike what the BSC model might suggest, we do take into account the fact that a faulty cell will remain faulty during the entire processing of one input.]}\seb{[A faulty cell will be used once and only once in the models used (no RNN).]}
We define $\bm{p}$ a vector of noise levels for each layer $\ell$ of the network. We are interested in optimizing $\bm{p}$ layerwise to provide an additional degree of freedom when searching for an optimal energy-accuracy trade-off. We can thus express the overall memory energy consumption of the network as
\begin{equation}\label{eq:consolayerwise}
    E(\bm{p}) = \zeta * \sum_\ell \eta(p_\ell) n_\ell \, , 
\end{equation}
where $n_\ell$ is the number of parameters in layer $\ell$ of the network, $\zeta$ the number of bits per parameter and $\eta(p_\ell)$ the energy per bit given in \eqref{eq:conso}. 

\section{Training low-energy networks}\label{sec:alg}
\subsection{Uniform noise approach}\label{sec:globalp}
A first approach for improving the energy efficiency of DNNs consists in assigning a common fault rate (or memory configuration) to all layers, and to train the network under this fault rate.
For the network forward pass, weights $\theta$ are sampled through the BSC to obtain $\tilde{\theta}$. For the backpropagation, real weights $\theta$ are updated with the gradient of the corrupted weights $\tilde{\theta}$. It is possible to manually tune the training noise $p_t$ to obtain a desired accuracy level.
% (francois: I removed this sentence because I don't understand it.) This approach provides a frontier for the accuracy-energy trade-off: we will later show than LaNMax improves this solution.

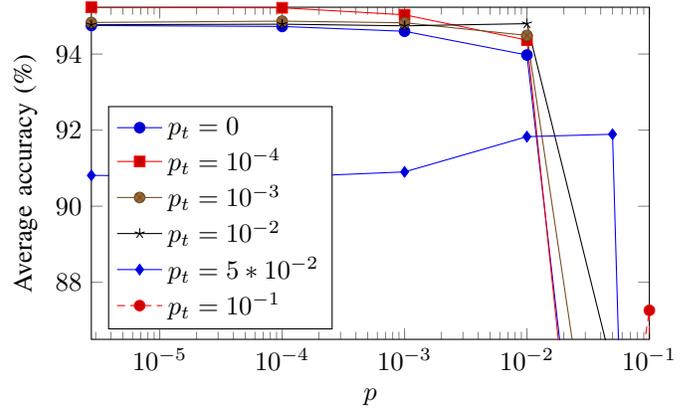
\begin{figure}
    \centering
    \pgfplotstableread[col sep = comma, skip first n=1]{saves/test_static_mean.csv}{\testmean}
    \begin{tikzpicture}
        \begin{axis}[ylabel=Average accuracy (\%), 
        xlabel=$p$, ymin=86.5, 
        legend pos=south west, 
        enlargelimits=false,
        xmode=log,
        height=6cm,
        width=9cm,
        legend cell align=left] %xticklabel={\pgfmathparse{100*e^\tick-0.01}\pgfmathprintnumber{\pgfmathresult}\%},cycle list name=black white]
            \foreach \n in {1, ..., 7} % -0.01: weird tick behavior with e^
            {
                \addplot+[mark size=2pt] table [x expr ={e^(-12.8*32*\thisrowno{8})}, y index=\n] {\testmean};
            }
            % \addplot+ coordinates {(0.009315625*32, 93.91)};
            % \addplot+ coordinates {(0.00380625*32, 84.27)};
            \legend{$p_t=0$, $p_t=10^{-4}$, $p_t=10^{-3}$, $p_t=10^{-2}$, $p_t=5*10^{-2}$, $p_t=10^{-1}$};%, $p=2*10^{-1}$};%, $p_{opt}$, $p_{\tau}$};
        \end{axis}
    \end{tikzpicture}
    \caption{Accuracy VS $p$ for a uniform training fault rate $p_t$.}
    \label{fig:etavsprec}
\end{figure}

% We make some observations about the uniform-p case that help guide the design of LaNMax:
Fig.~\ref{fig:etavsprec} shows the average accuracy achieved on CIFAR10 by networks trained at a fault rate $p_t$ but tested at a rate $p$.
We can observe a noise overfitting phenomenon when a network achieves the best test-set accuracy at $p=p_t$, which we observe when $p_t$ becomes sufficiently large. This effect was also reported in \cite{Hirtzlin2019OutstandingNetworks}.
Another interesting observation about the uniform noise case is that a small $p_t>0$ can provide a regularization that improves accuracy for all choices of $p$, which happens for $p_t=10^{-4}$ in this example.
It is important to keep in mind these phenomena when developing an automated approach for choosing $p_t$.
  
%In that case, training could reach local minima when decreasing the fault rate $p$, which may worsen the neural network loss.\seb{Not sure this one convey the good ideas. Pref. "training could reach local minima with $p_t$ such that decreasing the fault rate $p$ could worsen the neural network loss."} 
%However, we know this is an artefact since a network trained with lower $p_t$ can in fact achieve a better loss. 

%DONE expliquer la passe avant
%DONE checker theta: n_\ell a la place de theta quand on parle du nombre de paramètres
%DONE x values in decimal not exponential
%DONE bien préciser que le meilleur réseau avec une mémoire fiable est entraîné avec un petit bruit
%DONE préciser que l'on utilise le front de pareto par la suite
%DONE? marqueurs ouverts

\begin{figure}
    \centering
    \pgfplotstableread[col sep = comma, skip first n=1]{saves/df_plow.csv}{\plow}
    \begin{tikzpicture}
        \begin{axis}[boxplot/draw direction=y,
        boxplot/whisker range = 12,
        ylabel=Accuracy distribution (\%),
        xlabel=Layer index ($\ell$),
        grid=major,
        enlargelimits=false,
        height=5cm,
        width=9cm,]
        \foreach \n in {2, ..., 30} 
        {
            \addplot[boxplot] 
            table [y index=\n, row sep =\\] {\plow};
        }
        \draw[red] (0,92.6333348592122) -- (30,92.6333348592122);
        \end{axis}
    \end{tikzpicture}
    \caption{Sensitivity analysis under uniform noise $p=1\%$}
    \label{fig:pi}
\end{figure}
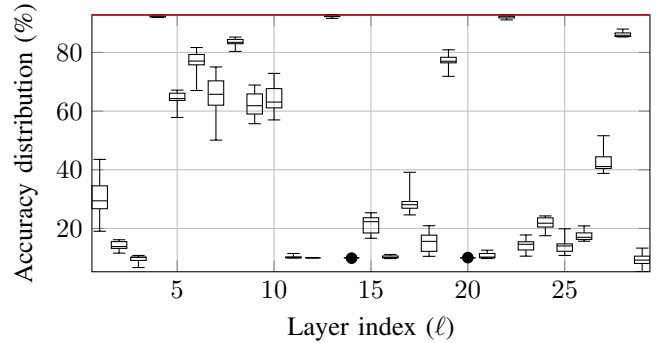

Starting from a uniform-noise solution that provides a good accuracy-energy trade-off, we can perform a sensitivity analysis to determine whether the reliability of some layers can be decreased further.
To do so, we tested the performance of the network when the parameters of layer $\ell$ are replaced with random values. As shown in Fig.~\ref{fig:pi}, some layers are critical for performance (accuracy close to 0 when randomized), but we see that some are less critical, motivating us to propose the LaNMax algorithm to systematically optimize the reliability of each layer.

%DONE bof 1 sur 2 libellé

% \begin{itemize}
%     \item natural configuration, tuning $p$ everywhere at once (manual tuning gives a rough frontier)
%     \item results obtain a local solution for $p$, but global optima require low p
%     \item sensitivity analysis: some layers aren't as necessary as others
%     \item need a method to fine tune per layer
% \end{itemize}

\subsection{LaNMax algorithm: finding the best layerwise $p$}
We model the accuracy-energy trade-off with an \emph{outer} loss function $L_O$ made up from the usual loss expectation $A$ and from the energy consumption $E(\bm{p})$. The usual loss expectation can be written as $A = \mathbb{E}(L_I(f(x|\bm{\tilde{\theta}}), y))$
with $\bm{\tilde{\theta}}$ the noisy weights, and $L_I(\cdot,\cdot)$ the \emph{inner} loss of the network output $f(x|\bm{\tilde{\theta}})$ for an input $x$ and a desired output $y$ (e.g. cross entropy).
Note that the loss expectation is taken over the inputs $x$, but also over the noisy weights $\tilde{\bm{\theta}}$.
We define the outer loss as 
\begin{equation}
 L_O(A, \bm{p}) = A + \alpha E(\bm{p}) + \lambda \bm{p} \, . 
\end{equation}
Parameter $\alpha$ allows us to set the desired trade-off between energy and accuracy: decreasing $\alpha$ will lead to solutions with better accuracy and larger energy.
Because of the possibility of noise overfitting mentioned in Sec.~\ref{sec:globalp}, it could be possible for an optimization scheme to remain stuck in a local minimum. A regularizer $\lambda \bm{p}$, which we call \emph{noise decay}, is used to correct this problem by adjusting $\lambda$.
% DONE fait de la fonction de perte habituelle L_I
% symbole grand A = E(L_I)
% ainsi que des composantes suivantes

We want to find the best reliability for each layer
\begin{equation}\label{eq:obj}
    \bm{p}^\star = \operatorname*{argmin}_{\bm{p}}    L_O(A, \bm{p}) \, ,
    % = & \operatorname*{argmin}_p \frac{1}{m} \sum_i L_I(f(x^{(i)};\bm{\tilde{\theta}}),y^{(i)}) + \alpha E(\bm{p}) + \lambda \bm{p} \, .
    % = & \operatorname*{argmin}_p A +\alpha E(\bm{p}) + \lambda \bm{p} \, . \nonumber
\end{equation}
subject to 
\begin{equation}
    0 < p_\ell \leq \frac{1}{2} \;\; \forall \ell \, .
\end{equation}
%\francois{I changed the left $\leq$ for $<$ since our energy model is not defined at $p=0$, and anyway it is not possible to reach zero error probability in a physical system.} OK
%$p$ is defined between those two extremum where there is no noise ($p=0$) and where the entropy is maximum, i.e. the weights could be random ($p=0.5$).
%

We propose to optimise $L_O$ with a Stochastic Gradient Descent \cite{bottou1998online} (SGD)-like algorithm (SGD$_O$).
Clearly, $L_I$ and $L_O$ cannot be optimized separately since $L_I$ depends indirectly on $\bm{p}$.
We thus need to adapt the network training so that $\theta$ and $\bm{p}$ are optimized jointly.
A possible approach would be to use Block Coordinate Descent\cite{Wright2015CoordinateAlgorithms} with alternating updates based on $L_I$ and $L_O$. Instead, to speed up the training process, we make use of the usual training mini-batch updates of $L_I$ to measure $L_O$ and approximate its gradient.

At each mini-batch, we apply a small random perturbation on $\bm{p}$: we thus obtain $\tilde{\bm{p}}$.
We can then first estimate $A$ conditioned on $\tilde{\bm{\theta}}$ for the current mini-batch (of size $m$), which yields $\hat{A}=\frac{1}{m} \sum_i  L_I(f(x^{(i)};\bm{\tilde{\theta}}),y^{(i)})$ and then estimate $L_O$. %\francois{[We could introduce $\hat{A}$ here? (see below)]}.\seb{[A is the expected value for $L_I$, and we only estimate/sample the random variable $L_I$ here (a realisation)]}
When all mini-batches have been seen for an epoch, a linear regression is performed on the estimates of $L_O$ of the current epoch, and the slope of this regression is used as an estimate of the outer gradient.
An additional advantage of this approach is that the randomness added to $\bm{p}$ reduces the likelihood of noise overfitting.
%The noise decay term of the loss function allows to further get the network out of a potential noise overfitting scenario.

To try and achieve a better convergence for SGD$_O$, we rely on a surrogate loss
\begin{equation}\label{eq:surr}
    \bar{L}_O(A,\bm{p}) = \frac{\hat{A}}{\hat{A}^\star} + \alpha \sqrt{E(\bm{p})} + \lambda \bm{p} \, ,
\end{equation}
where $\hat{A}^\star$ is the best loss observed on a mini-batch up to this point in the training of the neural network.
This formulation allows to stabilize the value of the first term (accuracy), which would vary due to the dynamics of the neural network training.
The gradient of the second term (energy) has large variations depending on the $p_\ell$ values. Taking the square root of the energy reduces these variations and we observe that this improves the SGD convergence.
% \begin{align}\label{eq:surr}
% \tilde{L}_O = & \frac{1}{m} \sum_i \frac{L_I(f(x^{(i)};\bm{\tilde{\theta}}))}{L_I^\star},y^{(i)}) +
%     \alpha \sqrt{\eta(\bm{p}}) + \lambda \bm{p} \, ,
% \end{align}
% DONE mieux présenter pourquoi first term

%DONE optimiser p et theta, les problemes ne sont pas séparables

The procedure to train a neural network using LaNMax is shown in Algorithm~\ref{alg:lanmax}. 
\begin{algorithm}[t]\label{alg:lanmax}
\caption{Layerwise Noise Maximisation (LaNMax) update}
\DontPrintSemicolon
% \LinesNumbered
\everypar={\nl}
% \KwIn{A training dataset of $M$ minibatches}
% \KwOut{$\bm{p}$ the fault rate for each layer $\ell$, $\bm{\theta}$ the NN parameters}
% \SetKwInOut{Require}{Require}
% \SetKwInOut{Hyperparameters}{Hyperparameters}
% \Require{Initial parameters $\bm{\theta}$ (network weights), $\bm{p}$ (memory fault rate), $M$ the number of mini-batches}
% \Hyperparameters{$s$: the epoch at which outer SGD is deactivated\\ $h$: a float to sample surroundings of $\bm{p}$\\ $\epsilon$: learning rate for both SGDs \\ Other SGD hyperparameters}
$K \leftarrow 1$ \\
\While{\label{line:epoch}stopping criterion not met}{
    \For{\label{line:minibatch}$k \in \{1,2,\ldots,M\}$}{
        Sample a minibatch $\{x^{(1)}, \ldots, x^{(m)} \}$ from the training set with targets $y^{(i)}$ \\ 
        \uIf{\label{line:s1} $K \leq s$}{
            $\delta_\ell \sim \mathcal{U}\{-h, 0, h\} \, \forall \ell$ \label{line:h}  \\
            $\tilde{\bm{p}}_k \gets \bm{p} + \bm{\delta}$ \label{line:randomized}\\
            $\tilde{\theta}_i \gets BSC(\theta_i, \tilde{p}_{k,\ell(i)}) \; \forall i$ \label{line:bsc} \\ %\tcp*{Where $\ell$ is the layer of parameter $i$}
            $L_k \leftarrow \bar{L}_O(\hat{A}, \tilde{\bm{p}}_k)$ \label{line:store}
     }
     \uElse{
        $\tilde{\theta}_i \gets BSC(\theta_i,p_\ell(i)) \; \forall i$ \label{line:usualloss}
     }
     $\bm{\hat{g}} \leftarrow \frac{1}{m} \nabla_\theta \sum_i L_I(f(x^{(i)};\bm{\tilde{\theta}}),y^{(i)})$ \label{line:usualgradient}\\%\tcp*{Gradient estimate of inner SGD}
     $\bm{\theta} \leftarrow \bm{\theta} - \epsilon \, \bm{\hat{g}}$\label{line:usualupdate} %\tcp*{Inner SGD update}
    }
    \uIf{\label{line:s2} $K \leq s$}{
        $\nabla_p L_O \leftarrow OLS_p(\{L_1, \ldots, L_M\})$\label{line:ols} \\%\tcp*{Gradient outer SGD}
        $\nabla_p L_O \leftarrow \frac{\nabla_p L_O}{||\nabla_p L_O||}$ \label{line:norma}\\ %\tcp*[h]{Gradient normalization}
        $\bm{p} \leftarrow \bm{p} - \epsilon \, \nabla_p L_O$ \label{line:update}%\tcp*{outer SGD update}
    }
    $K \leftarrow K+1$
}
\Return{$\bm{p}, \bm{\theta}$}
\end{algorithm}
%DONE remove comments and explain in text 
% \ell(i) instead of ...
% lanmax dans la forme de l'alg 1 est appliqué au sgd (adapted from SGD update rule in \cite{Goodfellow-et-al-2016}) mais on peut l'appliquer a d'autres variantes de sgd 
% DONE hyperparameter
Here, LaNMax is applied to a regular SGD training of a neural network's weights $\theta$ (adapted from the SGD update rule in \cite{Goodfellow-et-al-2016}) but it could be integrated to other SGD variants. The network is trained with the usual stopping criterion (line~\ref{line:epoch}), e.g. the number of epochs $K$. While iterating over the $M$ mini-batches (line~\ref{line:minibatch}) and if LaNMax is activated (line \ref{line:s1}), a random experiment is performed in the neighbourhood of $\bm{p}$ as follows. 
First, random perturbation amounts $\delta_\ell$ are sampled uniformly from the set $\{-h,0,h\}$ (line~\ref{line:h}). These values are assembled into a perturbation vector that is added to $\bm{p}$ (line~\ref{line:randomized}) in order to evaluate the loss function around $\bm{p}$. After sampling the random weights (line~\ref{line:bsc}), the loss function $L_O$ is computed and stored (line~\ref{line:store}). 
Repeating this experiment across all the mini-batches allows to iteratively refine the numeric gradient value.

If the network is training without LaNMax, the weights are instead sampled from the BSC at the fixed $\bm{p}$ (line \ref{line:usualloss}). The usual weight update for SGD$_I$ is performed either way at the end of a mini-batch (lines \ref{line:usualgradient}, \ref{line:usualupdate}). When all the mini-batches have been seen and if LaNMax is still active (line \ref{line:s2}), we perform SGD$_O$. First, a numerical gradient is computed via an ordinary least square (OLS) regression (line \ref{line:ols}). Then the gradient is normalized to ensure stable updates to $\bm{p}$ (line \ref{line:norma}). Finally, $\bm{p}$ is updated (line \ref{line:update}).

When training with LaNMax, it is advisable to stop updating $\bm{p}$ before the end of training through the parameter $s$ (lines \ref{line:s1} and \ref{line:s2}) so that the weights can be fine-tuned with the final $\bm{p}$ in the remaining training epochs. 
%It is possible to change the range of the random local changes with the parameter $h$ (line \ref{line:h}). 
Finally, as in a traditional SGD, a learning rate $\epsilon$ must be specified, but thanks to the gradient normalization in the SGD$_O$, it has a small impact and for all experiments that were performed it was sufficient to use the same learning rate for SGD$_I$ and SGD$_O$. 
%This allow to reduce the search space by removing an extra hyperparameter.

% If we're interested to converge to a finite set of $\bm{p} \in \mathcal{P}$ with $|\mathcal{P}|<|\bm{p}|$, for example in hardware implementations, it is possible to run a clustering algorithm at the epoch $s$. Hence, the $\bm{p}$ will be assigned to values in $\mathcal{P}$ for the remaining epochs.\francois{[But we do not do this in this paper, we should not give the idea away!]}

%%%%%%%%%%%
%%% REVIEW SUGGESTION : Why this network
%%%%%%%%%%%
\section{Experiments}\label{sec:exp}
We evaluate the energy reductions obtained by LaNMax by using it to train a WideResNet network~\cite{Zagoruyko2016WideNetworks} that has been shown to work well with binary weights \cite{McDonnell2018TrainingWeight} on the CIFAR10 dataset\cite{Krizhevsky2009LearningImages}.
As a point of comparison, we also evaluate the energy gains obtained by simply reducing the size of the network. 
For all results, we fix the number of layers to 28, and use the layer-width parameter of the WideResNet architecture to vary the number of parameters. We use a default width parameter of 10, and denote by $\rho$ the number of parameters normalized to this default network.
In addition, we also consider the simpler alternative of training the network with a uniform noise level applied to all layers.

% Energy model: choice of "a", and explanation that we consider that a reliable network has \eta=1 even though it is not defined in the model.
For all networks, the energy metric is defined by \eqref{eq:consolayerwise}. Note that \eqref{eq:conso} is not defined at $p=0$, since the fault probability can never be zero in a physical system. However, for simplicity, we consider that standard implementations have $p=0$, and an associated energy consumption $\eta(0) \triangleq 1$.
Finally, we set the model parameter $a=12.8$ as suggested in \cite{Hacene2019TrainingRobustness}.

% Choice of hyperparameters
For training, SGD$_I$ is configured as suggested in \cite{Zagoruyko2016WideNetworks}. Both SGD$_I$ and SGD$_O$ are configured with 200 epochs, a learning rate decayed every 60 epochs by a factor of 0.2 starting at 0.1, and a batch size of 128.
SGD$_I$ uses a Nesterov momentum of 0.9, a weight decay of $5*10^{-4}$ and a batch size of 128. SGD$_O$ uses a Nesterov momentum $\beta=0.2$ and a noise decay $\lambda=5\cdot 10^{-4}$.
The training set is augmented with padding, crops, and horizontal flips. The network weights are initialized following the state-of-the-art practices \cite{He2015DelvingClassification}.

% Accuracy measurement
When $p_\ell>0$, the network weights become stochastic at inference time, and therefore the network output is also stochastic.
To properly measure the average accuracy, we evaluate the test set multiple times, each time sampling new random weights, until the confidence interval on the average reaches a magnitude of five percents at 95\% confidence level.

% Design-space exploration with LaNMax (vary \alpha and \rho)
We perform several training runs using LaNMax while varying the accuracy-energy tradeoff parameter $\alpha \in \{0.1,0.05,0.03,0.02,0.01,0.001,0\}$, with LaNMax hyperparameters set to $h=0.01$ and $s=160$. Since we know from the uniform noise experiments that the network achieves a higher accuracy when $p=10^{-4}$ than when $p=0$, we also restrict $p \in [10^{-4}, 0.5]$, and $\bm{p}$ is initialized for all layers at $p_\ell=0.01$.

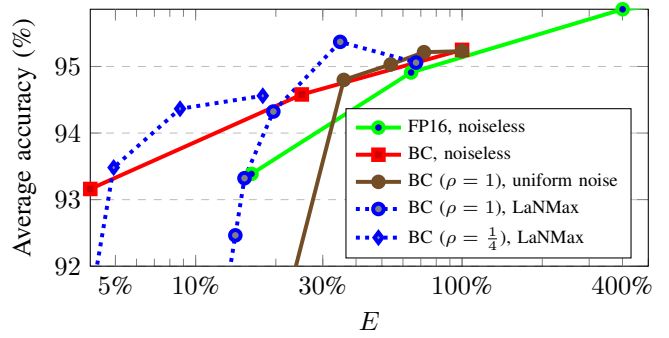
\begin{figure}
    \centering
    \begin{tikzpicture}
        \begin{axis}[ymajorgrids=true,
        grid style=dashed,
        xlabel={$E$},
        ylabel={Average accuracy (\%)},
        enlargelimits=false,
        xmode = log,
        legend pos=south east,
        ymin=92,
        xmax=5,
        legend style={font=\scriptsize},
        xticklabel={\pgfmathparse{round(100*e^\tick)}\pgfmathprintnumber{\pgfmathresult}\%},
        extra x ticks={0.05,0.3,4},
        height=5cm,
        width=9cm,
        legend cell align=left]
        % \addplot+ coordinates {(1,96.24) (9132634/36479194, 95.79) (1467610/36479194, 94.88) (369498/36479194, 93.34)}; %FP32
        \addplot+[line width=1.5pt, color=green] coordinates {(18239597/1139974.8125,96.07) (4566317/1139974.8125, 95.86) (733805/1139974.8125, 94.91) (184749/1139974.8125, 93.386)}; %FP16
        \addplot+[line width=1.5pt] coordinates {(1139974.8125/1139974.8125,95.25) (285394.8125/1139974.8125, 94.58) (45862.8125/1139974.8125, 93.16) (11546.8125/1139974.8125, 90.29)}; %BC
        %\addplot+[line width=1.5pt, mark=square, color=red, dotted,every mark/.append style={solid, fill=gray}] coordinates {(0.326809*1139974.8125/1139974.8125,95.25) (0.326809*285394.8125/1139974.8125, 94.58) (0.326809*45862.8125/1139974.8125, 93.16) (0.326809*11546.8125/1139974.8125, 90.29)};
        % min{-log(x)/(12.8 (1 + x log(2, x) + (1 - x) log(2, 1 - x)))}≈0.326809 at x≈0.0603608
        \addplot+[line width=1.5pt, mark size=2pt] coordinates {(1,95.23) (32*0.02248618254,95.21499999999999) (32*0.01686463691,95.03166666666665) (32*0.01124309127, 94.79833333333335) (32*0.00731379949, 91.89166666666667) (32*0.00562154563 ,87.26333333333334)  (32*0.00392929177, 47.59076923076923)}; % Faulty
        \addplot+[line width=1.5pt,mark size=2pt, mark=*, dotted,every mark/.append style={solid, fill=gray}, color=blue] coordinates {(0.09221365501753025,80.98250172932943) (0.11676303716278857,90.45166803995768) (0.14091579138510676,92.4633346557617)
            (0.1525395078722712,93.32166798909505)
            (0.19576898361529246,94.32333475748698) (0.3487740677788462,95.37000122070312) (0.6719246775977993,95.0583351135254)}; %LANMax
        \addplot+[line width=1.5pt,mark size=2pt,dotted,every mark/.append style={solid, fill=gray}, color=blue] coordinates {((0.1442193727264574)*(9123642/36461242)), 88.29000142415366) ((0.15903839403562522)*(9123642/36461242)), 91.30000178019206) ((0.19678900241021513)*(9123642/36461242)), 93.48166732788087) ((0.3493365938021792)*(9123642/36461242)), 94.36500091552735) ((0.7140793892991005)*(9123642/36461242)), 94.56000162760415)};
        %\addplot+ coordinates {((0.16265229020993127/32)*(367674/36461242)), 58.94185290866429) ((0.18398262491907674/32)*(367674/36461242)), 68.089488474528) (0.23302872251821244/32)*(367674/36461242)), 80.90714438302176) (0.36183956092338837/32)*(367674/36461242)), 87.19333445231119) (0.6073594659077954/32)*(367674/36461242)), 88.32500178019205)};
        \legend{
            {FP16, noiseless}, 
            {BC, noiseless}, 
            {BC ($\rho=1$), uniform noise}, 
            {BC ($\rho=1$), LaNMax}, 
            {BC ($\rho=\frac{1}{4}$), LaNMax}
            }; 
            %LaNMax($\rho=1\%$)}
        \end{axis}
    \end{tikzpicture}
    \caption{Average classification accuracy in terms of the normalized energy consumption for a network with 16-bit floating-point parameters and for BC networks with various noise configurations. For the FP16 and noiseless BC curves, energy reductions are obtained by reducing the number of parameters. For the other curves, the normalized number of parameters ($\rho$) is given.
    }
    \label{fig:thetalanmax}
\end{figure}

The accuracy results in terms of the energy metric are shown in Fig.~\ref{fig:thetalanmax}. The results obtained by a 16-bit floating-point (FP16) implementation are also shown as a point of comparison.
Different curves use different strategies for trading off accuracy for reduced energy.
For noiseless curves, energy is reduced by decreasing the number of parameters in the architecture. For the uniform noise BC, energy is reduced by increasing the unique $p$ parameter. Finally, for LaNMax results, the algorithm's $\alpha$ parameter is modified.
Note that for most data points of the uniform BC result, the test-time $p$ is the same as the training-time $p_t$, but since for $p=0$ it is better to train at $p_t=10^{-4}$, this result is shown instead.
% Observations:
% - uniform noise does not outperform noiseless BC by that much (how much?)
We can first see that a network trained with uniform noise for all layers provides a limited improvement in energy efficiency. For example the BC with uniform noise decreases energy by 28\% at an accuracy of 95.2\%, compared to the noiseless BC.
% - LaNMax provides gains over uniform noise
%%% gain BC uniforme / BC standard = 72% / 100% (@accuracy 95.21)
%%% gain lanmax / BC uniforme = ~30% / 72%  (@accuracy ~95.25)
%%% gain lanmax / BC standard = 30% / 100% (@accuracy ~95.25)
On the other hand, the more fine-grained optimization performed by LaNMax manages to find solutions with much smaller energy consumption at equal accuracy. At an accuracy of 95.3\%, the LaNMax solution decreases energy by $2.4\times$ with respect to the uniform-noise BC solution, and by $3.3\times$ with respect to the noiseless BC.

% - not many values of \alpha are in fact useful, we need to also reduce the network size to achieve good solutions at lower accuracies
It is interesting to note that for the energy-reliability model considered, 
changes in reliability alone do not allow to cover a wide range of accuracy values with good energy efficiency. Indeed, Fig.~\ref{fig:thetalanmax} shows that it is essential to also modify the number of parameters to move along the Pareto front.

\begin{figure}
    \centering
    \begin{tikzpicture}
    \begin{axis}[height=5cm, 
    legend pos=outer north east, 
    grid, ylabel=$p_\ell$, 
    xlabel=Layer index ($\ell$), 
    legend style={font=\scriptsize},
    legend cell align=left]
        \addplot+ coordinates {(1,1-0.9999) (2,1-0.9999) (3,1-0.9999)     (4,1-0.9999) (5,1-0.9999) (6,1-0.9999) (7,1-0.9999) (8,1-0.9999) (9,1-0.9999) (10,1-0.9999) (11,1-0.99909766) (12,1-0.98993598) (13,1-0.9999) (14,1-0.99003228) (15,1-0.98990838) (16,1-0.98992764) (17,1-0.9899004) (18,1-0.99006124) (19,1-0.98992507) (20,1-0.98872363) (21,1-0.9854627)  (22,1-0.9999) (23,1-0.98354119) (24,1-0.97976306) (25,1-0.98348715) (26,1-0.97624984) (27,1-0.98394924) (28,1-0.98206462) (29,1-0.9999)};
        \addplot+[mark=none] coordinates {(1,0.01) (29,0.01)};
        \legend{LaNMax $p_\ell$, Uniform noise $p=1\%$}
    \end{axis}
    \end{tikzpicture}
    \caption{The layerwise $p_\ell$ for the best result when varying $\alpha$ when $\rho=1$.} %TODO: It seems the caption is incorrect here? Isn't \alpha fixed at 0.001 as mentioned in the text? This will probably be mentioned by a reviewer, but in any case we should make sure to fix it in the final version.
    \label{fig:lanmaxstate}
\end{figure}
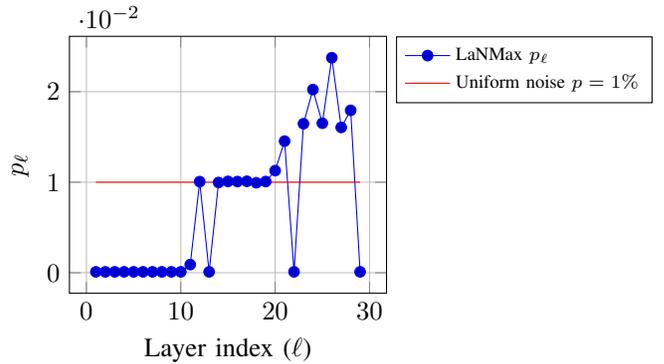

% - p-values vary significantly depending on the layer, we can explore the best solution found
Fig.~\ref{fig:lanmaxstate} shows the reliability assignment generated by LaNMax for $\rho=1$ and $\alpha=0.001$. This solution achieves an accuracy of 95.4\% for $E=35$\%.
As a point of comparison, the uniform-noise BC achieves 94.8\% for $E=36$\%, using $p=0.01$.
To improve the accuracy, the reliability of the first layers was increased to $p_\ell=10^{-4}$, while the reliability of the last layers was decreased. In the WideResNet architecture, deeper layers have more weights, making it more interesting to reduce the reliability of these layers.

%%%%%% OLD TEXT %%%%%%%
%\francois{Old text starts here.}

%We notice the shortcut convolutions (13th, 22th layers, and the 4th which blends in the adjacent layers) of the WideResNet architecture, which were robust in Fig.~\ref{fig:pi}, tend to have a low $p$ with LaNMax. 
%While this duality is surprising, it may be of interest to uncover the root causes. One potential explanation is that the relative energy cost of those shortcut may be quite low as they have fewer parameters than the other layers: hence, the outer SGD may have found little interest to reduce their reliability. While in the Fig.~\ref{fig:lanmaxstate}, the uniform $p$ noise seems in average higher, the few adaptations from LaNMax suffice to achieve a slightly lower energy for a higher accuracy.

\section{Conclusion}\label{sec:ccl}
% 1 - Contribution algorithmique
In this paper we proposed the LaNMax algorithm to optimise the storage energy of a neural network, with the possibility to control the energy-accuracy trade-off. 
We considered that the reliability of memory cells decreases exponentially as their energy usage is reduced.
LaNMax provides a way to automatically find the best weight-storage reliability for each layer of the neural network while retaining the accuracy.
This was achieved by adding an outer SGD-like optimization to the training, while re-using the results generated by the standard training algorithm to reduce overhead.

% 2 - Résumé résultats
For a binarized WideResNet on the CIFAR10 dataset, a LaNMax-optimized network achieves the same accuracy as the reliable network with more than three times less energy. 
%In the case of a uniformly faulty network, a saving of 2.4 times the memory energy can be seen. 
%This experiment results in the first layers to be highly reliable with an increasing fault rate deeper in the network.

% 3 - Interet general
The proposed training approach can help achieve ultra-low energy deep-learning inference by enabling implementations based on near-threshold CMOS circuits as well as emerging technologies.

% In this paper, we propose the LaNMax algorithm, an optimization scheme to minimize the storage energy consumption of a binarized neural network with a controlled trade-off between the final accuracy and the energy savings. By reducing the voltage supplied to the memory, we induce a fault rate $p$. We make use of LaNMax to find an adequate $p$ for each layer. In the general case, a three-fold reduction in energy consumption can be seen at isoaccuracy. LaNMax learns to distribute $p$ where needed: the solutions interestingly converge for a ResNet-like architecture to high reliability for the first layers, while the few last layers, with relatively larges amount of weights, have the lowest reliability. 

% With respect to the usual approaches, the proposed approach opens a new venue for energy optimization with the use of an outer SGD-like algorithm with a negligible overhead. Thus, we hope to contribute to a \emph{green AI} \cite{Schwartz2019GreenAI} at two levels: decreased resources needs at training and in production. To ensure the applicability of this idea in a general context, using LaNMax on a diverse set of task is still a necessary work. Future work will focus on expanding the energy savings in the context of LaNMax solutions.

% \begin{itemize}
%     \item lanmax allow to find with marginal computing an adequate layerwise noise, without any hypothesis on the impact of said noise
%     \item SGD in SGD goes well together, but requires slower movements on the outer loss function
%     \item future work will ...
% \end{itemize}

\section*{Acknowledgment}
The authors would like to thank IVADO, ReSMiQ and NSERC for their support.

\pagebreak
\bibliographystyle{./bibtex/IEEEtran_custom.bst}
\bibliography{./bibtex/IEEEabrv,references_custom}

\end{document}